\documentclass[times,twocolumn,final,authoryear]{elsarticle}

\usepackage{ycviu}
\usepackage{framed,multirow}

\usepackage{amssymb}
\usepackage{graphicx}

\usepackage{url}
\usepackage{cuted}
\usepackage{xcolor}
\definecolor{newcolor}{rgb}{.8,.349,.1}
\definecolor{Y}{RGB}{255, 179, 102}
\definecolor{X}{RGB}{130, 179, 102}
\definecolor{H}{RGB}{51, 153, 255}

\usepackage{pifont}
\newcommand{\cmark}{\ding{51}}%
\newcommand{\xmark}{\ding{55}}%

\usepackage{amsmath,amsfonts,bm}

\def\eqref#1{equation~\ref{#1}}

\def\1{\bm{1}}

\DeclareMathAlphabet{\mathsfit}{\encodingdefault}{\sfdefault}{m}{sl}
\SetMathAlphabet{\mathsfit}{bold}{\encodingdefault}{\sfdefault}{bx}{n}

\newcommand{\etens}[1]{\mathsfit{#1}}

\def\etK{{\etens{K}}}

\def\etQ{{\etens{Q}}}

\def\etV{{\etens{V}}}

\newcommand{\softmax}{\mathrm{softmax}}

\newcommand{\normltwo}{L^2}

\usepackage{url}
\usepackage{multirow}

\def\comment#1{{}}
\def\eg{{\em e.g.}}
\def\ie{{\em i.e.}}
\def\etal{{\em et al.}}

\journal{Computer Vision and Image Understanding}

\begin{document}

\ifpreprint
  \setcounter{page}{1}
\else
  \setcounter{page}{1}
\fi

\begin{frontmatter}

\title{Collaborative Three-Stream Transformers for Video Captioning}

\author[1]{Hao \snm{Wang}} 
\author[2]{Libo \snm{Zhang}\corref{cor1}}
\cortext[cor1]{Corresponding author:}
\ead{libo@iscas.ac.cn}
\author[3]{Heng \snm{Fan}}
\author[1]{Tiejian \snm{Luo}}

\address[1]{School of Computer Science and Technology, University of Chinese Academy of Sciences, Beijing, 101408, China}
\address[2]{Institute of Software, Chinese Academy of Sciences, Beijing, 100190, China}
\address[3]{Department of Computer Science and Engineering, University of North Texas, Denton 76203, Texas, United States of America}

\received{1 May 2013}
\finalform{10 May 2013}
\accepted{13 May 2013}
\availableonline{15 May 2013}
\communicated{S. Sarkar}

\begin{abstract}

As the most critical components in a sentence, subject, predicate and object require special attention in the video captioning task. To implement this idea, we design a novel framework, named COllaborative three-Stream Transformers (COST), to model the three parts separately and complement each other for better representation. Specifically, COST is formed by three branches of transformers to exploit the visual-linguistic interactions of different granularities in spatial-temporal domain between videos and text, detected objects and text, and actions and text. Meanwhile, we propose a cross-granularity attention module to align the interactions modeled by the three branches of transformers, then the three branches of transformers can support each other to exploit the most discriminative semantic information of different granularities for accurate predictions of captions. The whole model is trained in an end-to-end fashion. Extensive experiments conducted on three large-scale challenging datasets, \ie, YouCookII, ActivityNet Captions and MSVD, demonstrate that the proposed method performs favorably against the state-of-the-art methods.
\end{abstract}

\begin{keyword}
\MSC 41A05\sep 41A10\sep 65D05\sep 65D17
\KWD Video Captioning\sep Multi-Modal\sep Cross-Granularity\sep Spatial-Temporal Domain

\end{keyword}

\end{frontmatter}


\section{Introduction}
Video captioning aims to generate natural language descriptions of video content, which attracts much attention in recent years along with the rapidly increasing amount of videos recorded in daily life. It can be adopted for a wide range of real-world applications, such as blind people assistance, automatic videos summarization and classification, and intelligent video surveillance. However, as noted in former works \citep{DBLP:conf/eccv/XiongDL18,DBLP:conf/cvpr/ParkRDR19,DBLP:conf/acl/LeiWSYBB20}, it is very challenging to generate natural paragraph descriptions due to the difficulties of having relevant, less redundant, and semantic coherent sentences.

Recently, researchers attempt to use the transformer model to solve the video captioning task \citep{DBLP:conf/nips/VaswaniSPUJGKP17, DBLP:conf/acl/DaiYYCLS19,DBLP:conf/cvpr/IashinR20,DBLP:conf/cvpr/ZhuY20a,DBLP:conf/naacl/TangLB21}, which relies on the self-attention mechanism to describe the interactions between different modalities of the input data, such as video, audio, and text. In practice, the aforementioned methods generally concatenate the features extracted from individual modalities, or use self-attention to model the interactions between extracted features. Although they advance the state-of-the-art of video captioning, it is still far from satisfactory in real applications due to the domain gap between different modalities. Thus, a question naturally arises, ``\textit{How to reduce the domain gap and capture the interactions among visual and linguistic modalities for video captioning?}''

\begin{figure}
  \centering
  \includegraphics[width=\linewidth]{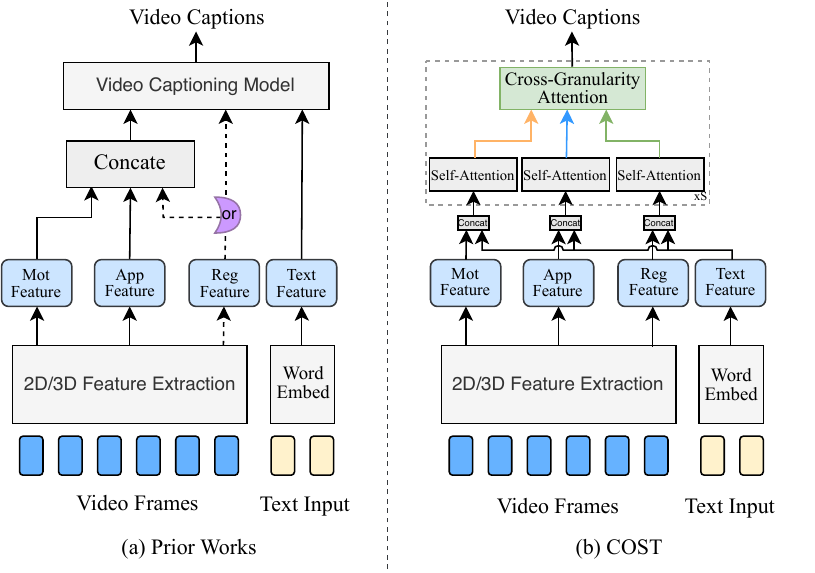}
  \caption{Comparison between prior works and the proposed COST. Previous works tended to \textbf{concatenate} the visual features, \ie, motion feature (Mot Feature), appearance feature (App Feature) and region feature (Reg Feature), as input to their video captioning model. Differently, we propose a three-branch transformer-based architecture to encode the visual-linguistic interactions of different granularities separately and design the cross-granularity module to complement the interactions with each other.}
  \label{fig:abstract}
\end{figure}
Before answering this question, let us see the basic grammar rules at first. As pointed out in \citet{DBLP:conf/iccv/KrishnaHRFN17} and \citet{DBLP:conf/aaai/ZhouXC18}, a sentence is generally presented as the following form, \eg, 
\begin{center}
\textit{\textcolor{red}{Women} \textcolor{green}{wear} \textcolor{blue}{Arabian skirts} on a stage}.
\end{center}
where \textcolor{red}{Subject}, \textcolor{blue}{Object}, and \textcolor{green}{Predicate} are the three most critical elements, and indicate the objects, the actions of objects, and the interactions among different objects, respectively. We believe that these components correspond to visual representations with different granularities, and modeling mutil-modal interactions based on them can effectively reduce the domain gap and help model understand the content of video better. Thus motivated, we propose a novel framework, called COllaborative three-Stream Transformers (COST), for video captioning. Different from former methods \citep{DBLP:conf/cvpr/ZhouZCSX18, DBLP:conf/acl/LeiWSYBB20, DBLP:conf/acl/DaiYYCLS19, DBLP:conf/iccv/WangZLZC021} which directly concatenate visual features of different granularities and model the correlation between visual and linguistic modality directly (Fig. \ref{fig:abstract}(a)), COST models the visual-linguistic \citep{DBLP:conf/cvpr/NanQXLLZL21, DBLP:conf/aaai/Fan020, DBLP:journals/corr/abs-2201-08264, DBLP:conf/cvpr/WeiZLZW20, DBLP:journals/corr/abs-2002-06353} interactions of different granularities separately in spatial-temporal domain and then fuses the multi-modal feature in different branches for better video content understanding (Fig. \ref{fig:abstract}(b)). As we argue that these previous works have overlooked the domain gap present in the features extracted from different pre-trained models, and this oversight renders it unreasonable to simply concatenate these features as a unified video representation. Note that, the dashed lines in Fig. \ref{fig:abstract}(a) indicate that there exist some works extracting and combining region features with the concatenated motion and appearance features \citep{DBLP:conf/cvpr/ParkRDR19} together or sending them parallel \citep{DBLP:conf/cvpr/ZhangSY0WHZ20} as the visual input to their models.

Specifically, the proposed COST consists of three transformer branches, including the Video-Text, Detection-Text, and Action-Text transformers. The Video-Text transformer is used to model the interactions between the global video appearances and linguistic texts, which makes the model perceive the general content of the video. The Detection-Text transformer aims at accurately locating objects in individual video frames, which enforces the model to focus on the objects being aligned in the visual and linguistic modalities, \ie, indicating the Subjects and Objects in caption sentences. The Action-Text transformer is designed to model the actions/relations of objects between the visual and linguistic modalities, \ie, indicating the Predicate in caption sentences. Meanwhile, to align the interactions modeled by the three branches of transformers, we introduce a cross-granularity attention model in COST. In particular, the similarity between interactions from different branches is computed to represent the relevance among visual modalities and help inject the information from other interactions. Note that we use the terminology "granularity" to describe the different levels of detail or scales at which the same input data is processed. For example, our motion features and region features (detection features) are both extracted from images, yet they provide distinct information. In contrast, "modality" is a related term that refers to the various types of inputs that a machine learning system processes, such as text, image, and audio. Each modality requires specific processing techniques and algorithms to extract valuable information and patterns effectively. In addition, we introduce additional training objectives for Detection-Text and Action-Text streams separately to align the semantics of embeddings to the underlying video information which is supposed to be conveyed at the time of feature extraction, instead of only setting cross-entropy loss between generated captions and ground-truth descriptions like most former methods. As we believe that introducing appropriate guidance can ensure the uniqueness of information in each stream and provide complementary information for each other. In this way, different branches of transformers support each other to exploit more discriminative semantic information in different modalities and granularities, and enforce the model to pay more attention on generating the accurate Subject, Object and Predicate predictions. The whole model is trained in an end-to-end fashion using Adam algorithm \citep{DBLP:journals/corr/KingmaB14}.

Extensive experiments are conducted on three publicly challenging datasets, \ie, YouCookII \citep{DBLP:conf/aaai/ZhouXC18}, ActivityNet Captions \citep{DBLP:conf/iccv/KrishnaHRFN17} and MSVD \citep{DBLP:conf/acl/ChenD11}, to demonstrate the superior performance of the proposed method compared to the state-of-the-art methods \citep{DBLP:conf/cvpr/ZhouZCSX18,DBLP:conf/acl/DaiYYCLS19,DBLP:conf/cvpr/ParkRDR19,DBLP:conf/acl/LeiWSYBB20, DBLP:conf/iccv/WangZLZC021, DBLP:journals/corr/abs-2201-08264, DBLP:journals/corr/abs-2111-13196}. Specifically, our method achieves very competitive CIDEr scores with TSN \citep{DBLP:journals/pami/0002X00LTG19} features as input, \ie, $45.54\%$ and $24.77\%$, on the YouCookII \textit{val} set and the ActivityNet \textit{ae-test} set,  surpasses or approaches the state-of-the-arts.

The main contributions of this paper are summarized as follows:

\begin{enumerate}
    \item We propose a novel framework, \ie, COllaborative three-Stream Transformers (COST), that leverages multiple transformer branches to explore various components in a sentence for video captioning.

    \item We design a simple but effective cross-granularity attention module to align the interactions modeled by different transformer branches, which supports each other to exploit discriminative semantic cue from different granularities for more accurate predictions of captions.

    \item We specially introduce a new training objective for COST that constrains the semantics of embeddings in Detection-Text branch and Action-Text branch to supply  for caption generation, further enhancing performance.

    \item Extensive experiments conducted on three challenging datasets show that our method performs favorably against the state-of-the-art methods.
\end{enumerate}

\section{Related Work}
{\noindent {\bf Video Captioning.}} Over the past few years, video captioning has received increasing attention from both computer vision and natural language processing community \citep{xu2021artificial}, which aims to generate linguistic description for the video content. Early works mainly focus on template-based approaches. For example, \citet{DBLP:conf/iccv/GuadarramaKMVMDS13} and \citet{DBLP:conf/cvpr/DasXDC13} first detect the visual objects in a video with human-crafted features and then use them to fill the pre-defined templates with slots. However, such methods are restricted in generating semantically-rich sentences due to the high dependence on fixed templates and language rules. 

Motivated by the success of neural network in translation task \citep{DBLP:conf/nips/SutskeverVL14}, the methods of taking captioning task as translation task became popular \citep{DBLP:conf/naacl/VenugopalanXDRM15, DBLP:conf/cvpr/YuWHYX16, DBLP:conf/cvpr/PanYLM17, DBLP:conf/eccv/ZhangHS18}. The nature of these methods is performing sequence-to-sequence learning in an encoder-decoder paradigm \citep{DBLP:conf/ijcai/ChenYJ19}, where the convolutional neural networks (CNNs) \citep{DBLP:conf/cvpr/ZhengWT20} and Long-Short Term Memory (LSTM) networks \citep{DBLP:conf/cvpr/PeiZWKST19,DBLP:conf/cvpr/ParkRDR19} are generally adopted to extract discriminative feature embeddings from input video and generate accurate captions separately. To alleviate the heavy computational burden of applying 2D-CNN, especially 3D-CNN to dense frame inputs for visual feature extraction, the methods in this paradigm mostly operate on the pre-extracted features and the subsequent works mainly concentrate on the improvements in feature extraction or utilization, including 1) multi-modal feature extraction \citep{DBLP:conf/naacl/WangWW18, DBLP:conf/aaai/HaoZG18a, DBLP:conf/cvpr/HoriHWWLCM18, DBLP:journals/tcsv/XuLWZNSK19, DBLP:conf/cvpr/ZhangSY0WHZ20} and 2) feature utilization optimization \citep{DBLP:conf/naacl/VenugopalanXDRM15, DBLP:conf/iccv/VenugopalanRDMD15, DBLP:conf/iccv/YaoTCBPLC15, DBLP:conf/ijcai/LiZL17, DBLP:journals/pami/GaoLSS20, DBLP:conf/aaai/ChenJ19}. In the former one, HACA \citep{DBLP:conf/naacl/WangWW18} proposes a hierarchically LSTM-based network to learn and align the attentive representations of both visual and audio features at different granularities. DS-RNN \citep{DBLP:journals/tcsv/XuLWZNSK19} proposes a dual-stream framework to model visual and semantic features independently and decode the hidden states from both modalities jointly. ORG-TRL \citep{DBLP:conf/cvpr/ZhangSY0WHZ20} not only proposes an object relational graph to connect each object in video and do relational reasoning by graph convolutional network, but also designs a teacher-recommended learning method to utilize the external language model to improve the generation of caption model by integrating the linguistic knowledge. In the later one, LSTM-YT \citep{DBLP:conf/naacl/VenugopalanXDRM15} proposes that mean pooling features across all frames is a rational representation for generating simple descriptions to short video clips. However, it totally ignores the order of frames in original video and discards the temporal information which is crucial for video captioning. S2VT \citep{DBLP:conf/iccv/VenugopalanRDMD15} firstly introduces sequence-to-sequence model to video captioning task and processes frames and generates words sequentially while preserving the temporal structure. In order to make the model concentrate on relevant features for specific word generation, temporal attention \citep{DBLP:conf/iccv/YaoTCBPLC15} is proposed to assign higher weights to relevant features. Apart from the frame-level attention, MAM-RNN \citep{DBLP:conf/ijcai/LiZL17} proposes that different regions in the video frame contribute differently to the word prediction, and designs a two-layers structure with the first layer focusing on the most salient regions in each frame and the second one attending to the most correlated frames. 

Recently, inspired by the superior performance in learning long-range relations through the attention mechanism compared with RNNs(\eg, LSTM and GRU \citep{DBLP:journals/corr/ChungGCB14}), Transformer-based models have emerged in this field. \citet{DBLP:conf/cvpr/ZhouZCSX18} proposes an end-to-end trained transformer model, where the encoder is designed to extract semantic representations from the video, and the proposal decoder receives the encoding output with different anchors to form video event proposals. \citet{DBLP:conf/iccv/SunMV0S19} designs the VideoBERT model to learn bidirectional joint distributions over sequences of visual and linguistic tokens. \citet{DBLP:conf/acl/LeiWSYBB20} develops the memory-augmented recurrent transformer, which uses a highly summarized memory state from the video clips and the sentence history to facilitate better prediction of the next sentence. More recently, by employing the tubelet embedding scheme and factorized encoder architecture from ViViT \citep{DBLP:conf/iccv/Arnab0H0LS21} and elaborately designed bi-directional objective, MV-GPT \citep{DBLP:journals/corr/abs-2201-08264} can be applied to raw pixels instead pre-extracted visual features and generates captions directly. SwinBERT \citep{DBLP:journals/corr/abs-2111-13196} is also an end-to-end transformer-based video captioning model with video frame patches as input, which uses VidSwin \citep{DBLP:journals/corr/abs-2106-13230} as visual encoder to extract video tokens and promotes cross-modality representation by applying masked text token prediction. They provide new insights for the future development of this field.

{\noindent {\bf Multi-modal cross-attention mechanism.}} The interactions between different modalities are critical for the video captioning task.  Recent transformer based methods \citep{DBLP:conf/cvpr/IashinR20,DBLP:conf/cvpr/ZhuY20a,DBLP:conf/naacl/TangLB21} use the cross-attention module to learn correlations across different modalities. For example, \citet{DBLP:conf/cvpr/IashinR20} concatenates the learned embeddings from multiple modalities, \eg, video, audio and speech, for event description. \citet{DBLP:conf/naacl/TangLB21} uses frame-level dense captions as an auxiliary text input for better video and language associations, where the constrained attention loss is used to force the model to automatically focus on the best matched caption from a pool of misalignment caption candidates. Recently, \citet{DBLP:conf/aaai/Li0022} proposed a novel approach to strengthen image and text embeddings through the incorporation of action information through cross-attention manner, and achieves state-of-art performance on image-text retrieval task.

Despite sharing some similarities with these approaches in using self-attention to align visual-linguistic interactions, our approach is significantly different. In specific, we design the cross-granularity attention module integrated in the collaborative three-stream transformers to align three types of visual-linguistic interactions of different granularities, as we believe that they could support and complement each other, for example, the model could more easily understand the current motion scene after injecting the information from detection feature of "basketball" and action feature of "play" to video feature, leading to more discriminative semantic cues for better caption generation.

\noindent {\bf Multi-branch architectures.} The idea of multi-branch has been widely applied in various tasks \citep{DBLP:journals/pami/LiuRY21, DBLP:conf/cvpr/ZhuY20a, DBLP:journals/corr/abs-2010-03855}. SibNet \citep{DBLP:journals/pami/LiuRY21} proposes a two-branch architecture to encode the content and semantic of videos separately and then the both branches are combined and fed to decoder for video caption generation. ActBert \citep{DBLP:conf/cvpr/ZhuY20a} proposes the tangled transformer block to encode three sources of information, \ie, global actions, local regional objects, and linguistic descriptions to learn better video-text representation. MTTSNet \citep{DBLP:journals/corr/abs-2010-03855} defines three region features corresponding to part-of-speech (POS) relation and designs a triple-stream network to encode these features separately and merges the processed embeddings to generate caption and predicts POS of each word for image caption task. 

Our COST also adopts the multi-branch architecture, but it is different than the above methods from two aspects. First, the design motivations are different. We use three-branch architecture to encode the visual-linguistic interaction of different granularities to disentangle the complex video information by self-attention, and use the cross-granularity attention module to complement and enhance the embeddings extracted from different branches, so as to improve the video analysis ability of the model to generate accurate description. Second, the contributions of branch to task are different. For example, ActBert and MTTSNet only set optimization objective for the whole model instead of each branch, and SibNet designs specific loss for two branches but only to improve the ability of the model to extract feature from the video. Differing from them, we set training objective for each branch to align the semantics of pre-extracted visual features to correspond to POS, and the embeddings in each branch are semantically complementary and directly participate in caption generation after information fusion. 

{\noindent {\bf Multi-modal pre-training models.}} Large-scale pre-training is another effective way to improve the accuracy of captioning models. Specifically, the jointly trained video and language models \citep{DBLP:conf/iccv/SunMV0S19, DBLP:conf/cvpr/ZhuY20a, DBLP:conf/ijcnlp/HuangPZRS20,DBLP:journals/corr/abs-2002-06353,DBLP:conf/nips/GingZPB20, DBLP:journals/corr/abs-2201-08264} on the large-scale datasets, such as YouTube-8M \citep{DBLP:journals/corr/Abu-El-HaijaKLN16} and HowTo100M \citep{DBLP:conf/iccv/MiechZATLS19} with automatic speech recognition\interfootnotelinepenalty=10000 \footnote{\url{https://developers.google.com/youtube/v3/docs/captions}} transcripts, provide discriminative features for downstream tasks, such as video captioning, action localization and \etal. VideoBert \citep{DBLP:conf/iccv/SunMV0S19} and ActBert \citep{DBLP:conf/cvpr/ZhuY20a} collect large paired video sequences and text descriptions with the help of off-the-shelf automatic speech recognition (ASR) model, and construct the BERT-style objective to train the video and text encoders simultaneously. \citet{DBLP:conf/ijcnlp/HuangPZRS20} constructs a dense video captioning dataset, \ie, Video Timeline Tags (ViTT), and explores several multi-modal sequence-to-sequence pre-raining strategies using transformers \citep{DBLP:conf/nips/VaswaniSPUJGKP17}. \citet{DBLP:journals/corr/abs-2002-06353} also applies transformers with two single-modal encoders to encode the video and text separately, a cross encoder to model the interactions between video and text representations, and a decoder to reconstruct or generate text. \citet{DBLP:conf/nips/GingZPB20} develops the Cooperative hierarchical Transformer (COOT) to model the interactions between different levels of granularities and modalities, which achieves superior results on video-text retrieval task and provides learned representations to improve the performance of video captioning model \citep{DBLP:conf/acl/LeiWSYBB20} significantly. Recently, \citet{DBLP:journals/corr/abs-2201-08264} proposes that recent visual-language pre-training frameworks lack the ability to generate sentences and presents MV-GPT with novel bidirectional objective for generation task.

\section{Our Approach}
As discussed above, we design the collaborative three-stream transformers to model the interactions of objects, and actions/relations of objects between different modalities, which is formed by three branches of transformers, \ie,  Video-Text, Detection-Text, and Action-Text transformers. Specifically, the video and text inputs are firstly encoded to extract the multi-modal feature embeddings. After that, the embeddings are fed into the three-stream transformers to exploit the interactions between 
linguistic embedding and visual embeddings of different granularities in spatial-temporal domain, \ie, global videos, object regions, and actions. Meanwhile, the cross-granularity attention module is designed to align the interactions modeled by the three branches of transformers. The overall architecture of the proposed method is shown in Fig. \ref{fig:structure}.

\subsection{Multi-Modality Tokens}
Three kinds of tokens, \ie, visual tokens, linguistic tokens, and special tokens, are used to express the video and text inputs, which are described as follows. 

\begin{figure*}[ht]
  \centering
  \includegraphics[width=0.96\linewidth]{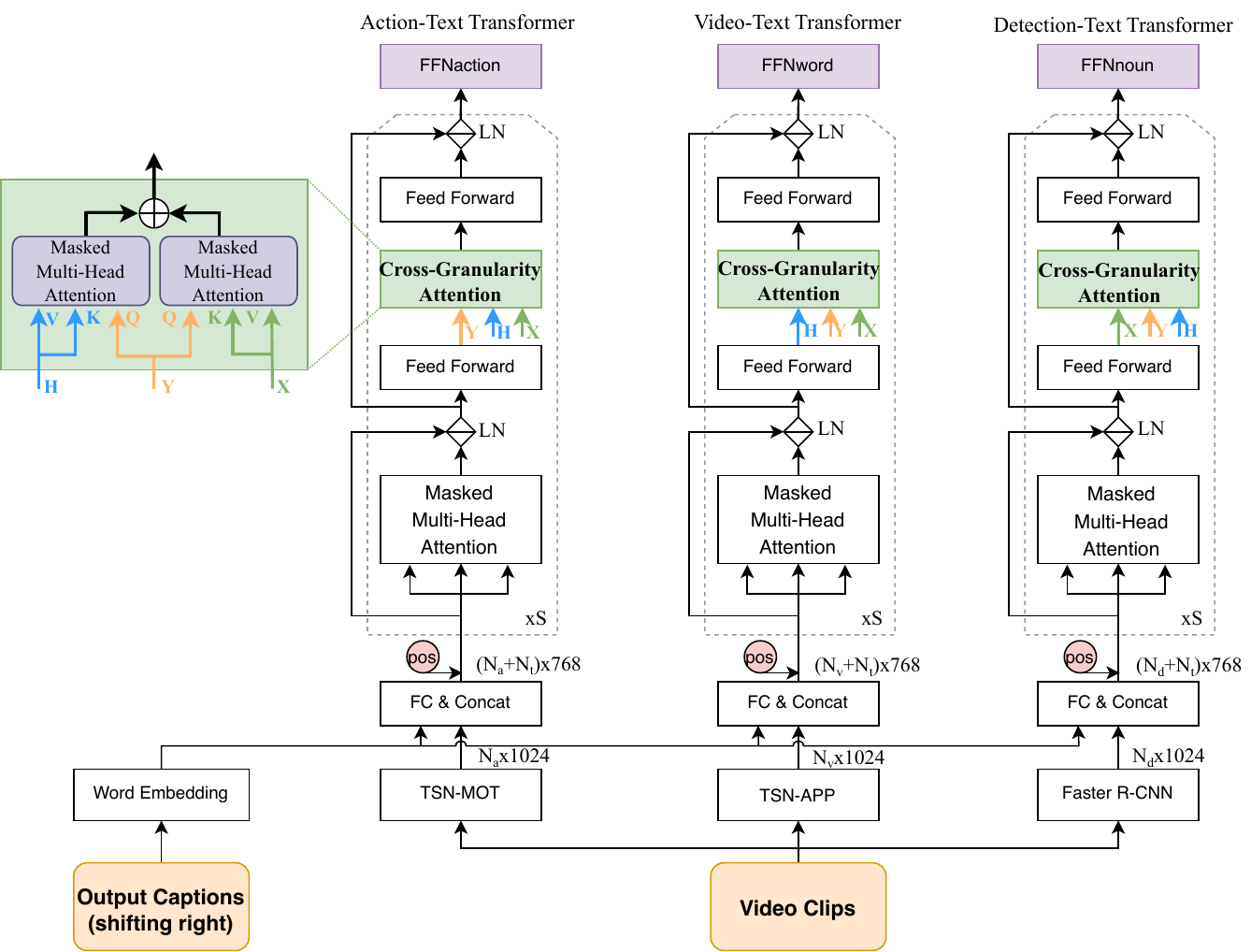}
  \caption{The network architecture of the proposed COST method, which is formed by three branches of transformers, \ie, the Action-Text, Video-Text and Detection-Text transformers. The cross-granularity attention module is designed to align the interactions modeled by the three different branches of transformers, and \textcolor{Y}{Y}, \textcolor{H}{H} and \textcolor{X}{X} represent the interactions of these branches separately.}
  \label{fig:structure}
\end{figure*}

{\noindent {\bf Visual tokens.}}
For the visual tokens, we use three kinds of tokens for different granularities in spatial-temporal domain, that is the video tokens, the detection tokens, and the action tokens. 
\begin{itemize}
\item \textit{Video tokens} provide the global semantic information in the video sequence. In contrast to \citet{DBLP:conf/acl/LeiWSYBB20}, we only use the appearance features extracted by Temporal Segment Networks (TSN) \citep{DBLP:conf/eccv/WangXW0LTG16} (denoted as TSN-APP in Fig. \ref{fig:structure}) as the video tokens, \ie, $\{f^v_1, f^v_2, \cdots, f^v_{N_v}\}$, where $f^v_i$ is the extracted feature of the $i$-th video clip, and $N_v$ is the number of video clips. Notably, we can also leverage more powerful multi-modal feature extraction method COOT \citep{DBLP:conf/nips/GingZPB20} to improve the performance, which is pre-trained on the large-scale HowTo100M dataset \citep{DBLP:conf/iccv/MiechZATLS19}.   

\item \textit{Detection tokens} are used to enforce the model to focus on the Subjects or Objects in caption sentences. Similar to \citet{DBLP:conf/cvpr/ParkRDR19,DBLP:conf/nips/LuBPL19,DBLP:conf/cvpr/ZhuY20a}, we use the Faster R-CNN method with 101-layer residual net (ResNet-101) \citep{DBLP:conf/cvpr/HeZRS16} as backbone to detect the objects in each frame, which is pre-trained on the Visual Genome dataset \citep{DBLP:journals/ijcv/KrishnaZGJHKCKL17}. After that, the detection features in Faster R-CNN corresponding to the objects with the highest confidence scores in $K$ categories\footnote{If the category number of the detected objects is less than $K$ in a frame, we select the $K$ detected objects with the highest confidence scores regardless the object categories to generate the detection tokens.}  are used to generate the detection tokens for each frame. We use $\{f^d_1, f^d_2, \cdots, f^d_{N_d}\}$ to denote the set of detection tokens, where $f^d_i$ is the $i$-th detection feature, and $N_d$ is the total number of detections in the video sequence.

\item \textit{Action tokens} are designed to enforce the model to concentrate on the Predicates in caption sentences. Following \citet{DBLP:conf/acl/LeiWSYBB20}, the optical flow features of video sequences are extracted by TSN \citep{DBLP:conf/eccv/WangXW0LTG16} (denoted as TSN-MOT is Fig. \ref{fig:structure}) to generate the action tokens, which are used to describe the actions/relations of objects. The action tokens are denoted as $\{f^a_1, f^a_2, \cdots, f^a_{N_a}\}$, where $f^a_i$ is the motion feature of the $i$-th video clip, and $N_a$ is the total number of action tokens. It is noted that the $N_v$ is always equal to $N_a$ in our method based on the pre-trained model we use for extracting appearance features and motion features.
\end{itemize}

{\noindent {\bf Linguistic tokens.}}
We break down the captions of video sequences into individual words and compute the corresponding linguistic tokens using the GloVe model \citep{DBLP:conf/emnlp/PenningtonSM14}. The linguistic tokens are denoted as $\{f^t_1, f^t_2, \cdots, f^t_{N_t}\}$, where $f^t_i$ is the extracted features of the $i$-th word using the GloVe model, and $N_t$ is the total number of words. In light of the emergence of more powerful language representation models, such as BERT \citep{DBLP:conf/naacl/DevlinCLT19} and CLIP \citep{DBLP:conf/icml/RadfordKHRGASAM21}, GloVe appears outdated as a model for vector representations for text. Nonetheless, we utilize GloVe features to represent captions for the sake of comparison fairness, as the main methods under comparison, such as MART \cite{DBLP:conf/acl/LeiWSYBB20} and PDVC \cite{DBLP:conf/iccv/WangZLZC021}, rely on GloVe as their text encoding model. Additionally, setting feature vectors to special characters by first random initialization with the same shape as features provided by GloVe, such as "[EOS]", and the semantics of these characters can to be learned during the training process, rendering GloVe a suitable choice for the current task.

{\noindent {\bf Special tokens.}} 
Besides the aforementioned tokens, we also introduce three kinds of special tokens in transformer, similar to BERT \citep{DBLP:conf/naacl/DevlinCLT19}. The first one is the granularity token $[\text{CLS}]$, which is added at the beginning of visual features to denote which granularity the following tokens belong to. The second one is the three kinds of separation token, \ie, $[\text{SEP}]$, $[\text{BOS}]$, and $[\text{EOS}]$. $[\text{SEP}]$ is used at the end of the visual tokens to separate them from the linguistic tokens, $[\text{BOS}]$ is used to denote the beginning of linguistic tokens, and $[\text{EOS}]$ is used to denote the ending of the linguistic tokens, respectively. And the last one is the padding token $[\text{PAD}]$ with two purposes: supplementing the visual token sequence or text token sequence to specified length for training or inference in parallel; substituting for all linguistic tokens in inference phase so no caption information will be leaked to the model. In addition, we use a fully-connected layer to encode the aforementioned tokens to the same dimension. Thus, the inputs for the three-stream transformer are computed as
\begin{small} 
\begin{equation}
	\begin{split}
\big\{[& \text{CLS}^{(\cdot)}], f^{(\cdot)}_1, f^{(\cdot)}_2, \cdots, f^{(\cdot)}_{N_{(\cdot)}},{[\text{SEP}]}, [\text{PAD}]_{\times N_1}, \\
	&[\text{BOS}], f^t_1, \cdots, f^t_{N_t}, [\text{EOS}], [\text{PAD}]_{\times N_2}\big\}
	\end{split}
\end{equation}
\end{small}
where $(\cdot)\in\{v, d, a\}$ indicates the video, detection, and action tokens, respectively. And the subscript of [\text{PAD}] means the number of it to supplement. Please note that the purpose of concatenating visual and linguistic tokens is not to directly fuse their information. Rather, the interaction between the features of two modalities will be encoded through a subsequent masked multi-head attention module. Thus, we choose concatenation as it helps to preserve the information of both modalities. We also use the positional encoding strategy \citep{DBLP:conf/nips/VaswaniSPUJGKP17} in the Video-Text, Detection-Text, and Action-Text transformers to describe the order information of caption sentences.

\subsection{Three-Stream Transformers}
As shown in Fig. \ref{fig:structure}, we feed the aforementioned tokens into the three-stream transformers. The Video-Text, Detection-Text, and Action-Text branches are formed by $S$ basic blocks, and each block mainly consists of a self-attention module and a cross-granularity attention module. Both the self-attention and cross-granularity modules are followed by a feed forward layer.

{\noindent {\bf Self-attention module.}} 
The self-attention module is designed to model the visual-linguistic alignments in each branch of transformers, \ie, Video-Text, Detection-Text, and Action-Text. Following \citet{DBLP:conf/nips/VaswaniSPUJGKP17}, we compute the attention function between different tokens as follows.  
\begin{equation}
\mathcal{A}(\etQ,\etK, \etV) = \softmax\big( \frac{\etQ{\etK}^\mathrm{T}}{\sqrt{d}} \big) \etV
\end{equation}
where $\etQ$, $\etK$ and $\etV$ are created by inputting tokens into three different fully-connected layers in each branch. And their dimensions are $\mathbb{R}^{N\times d}$, where $N$ and $d$ are the number of tokens and the dimension of embeddings, respectively. We advocate $h$ paralleled heads of scaled dot-product attentions to increase the diversity. It is noted that we also add mask to the scaled dot-product before applying Softmax to prevent the model from seeing future words \citep{DBLP:conf/nips/VaswaniSPUJGKP17}. We denote the whole module as \textit{Masked Multi-Head Self-Attention} in Fig. \ref{fig:structure} and the output is merged with input using a residual connection and layer norm.

\begin{figure}
  \centering
  \includegraphics[width=0.95\linewidth]{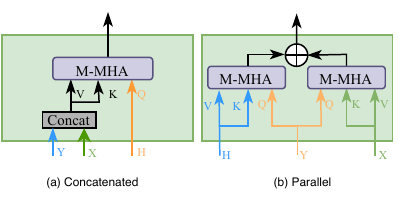}
  \caption{Two cross-granularity attention architectures to merge embeddings from different branches. M-MHA denotes the Masked Multi-Head Attention. \textcolor{X}{X}, \textcolor{Y}{Y} and \textcolor{H}{H} indicate the hidden states from Detection-Text branch, Action-Text branch and Video-Text branch separately and both sub-figures demonstrate the process of information fusion from the other two branches to Action-Text branch.}
  \label{fig:cross_attention_arch}
\end{figure}

{\noindent {\bf Cross-granularity attention module.}} 
Besides the self-attention module, we use the cross-granularity attention module to align the interactions modeled by the three branches of transformers. Specifically, we complement and enhance the embeddings in each branch by injecting the information from other branches based on cross-attention between them. The branch that is incorporated into information and the other two branches are dubbed as injected branch and supply branches for better understanding the following description. We propose two architectures as shown in Fig. \ref{fig:cross_attention_arch} and compare their performance by experiments. It is important to note that we only illustrate the implementation of the cross-granularity attention module within the Action-Text branch in Fig \ref{fig:cross_attention_arch}. Specifically, embeddings from this branch are utilized as the query to fuse information from the other two branches, while the embeddings from the other two branches similarly serve as queries within their respective cross-granularity attention modules.

\noindent \textit{Concatenated Architecture} A simple and straightforward way to inject the related information from supply branches is to concatenate the embeddings from them and use the obtained result as key and value while regarding the embeddings from injected branch as query in the multi-head attention layer. The whole process in Action-Text branch is shown in Fig. \ref{fig:cross_attention_arch}(a) and the other two branches perform the same computation with their own embeddings as query. The computation process is
\begin{gather}
\mathsf{M}_{\mathsf{Y}} = \mathrm{softmax}\big(\mathsf{Y} \odot \big( \bigl[\mathsf{H};\mathsf{X}\bigr]^\mathrm{T}\big)\big) \\
\mathsf{Y}' = \mathrm{FFN}\big(\mathsf{M}_{\mathsf{Y}}\odot\big(\bigl[\mathsf{H};\mathsf{X}\bigr]\big)\big)
\label{equ:concate}
\end{gather}

where $\odot$ represents the dot product, $\bigl[\cdot;\cdot\bigr]$ denotes the concatenation operation, $\mathrm{FFN}$ is the feed forward layer, $\mathsf{X}, \mathsf{Y}$ and $\mathsf{H}$ denote the hidden states from Detection-Text branch, Action-Text branch and Video-Text branch separately. We estimate the modality-wise normalization score of the interactions using a softmax layer. It should be noted that we add the mask here to prevent possible linguistic information leakage to the model. The affinity matrix $\mathsf{M}_{\mathsf{Y}}\in\mathbb{R}^{(N_a+N_t)\times (N_v+N_d+2N_t)}$ and $\mathsf{M}_{\mathsf{Y}}(i,j)$ denotes the normalized interaction score between the $i$-th entity in the Action-Text embeddings and the $j$-th entity in the concatenated Video-Text and Detection-Text embeddings. Based on the matrix, the feature embeddings of the Action-Text transformer can extract information from other branches. Feed forward layer is used to further encode the extracted information which will be merged to Action-Text transformer.

\textit{Parallel Architecture} As shown in \ref{fig:cross_attention_arch} (b), the other architecture design of cross-granularity attention is to feed embedding from one of supply branches and that from the injected branch to multi-head attention layer parallel, and then fuse the result from the two attention layers. The computation process is

\begin{gather}
\mathsf{M}_{\mathsf{YH}} = \mathsf{softmax}(\mathsf{Y} \odot \mathsf{H}^\mathrm{T}) \;\;\;\;\; \mathsf{M}_{\mathsf{YX}} = \mathrm{softmax}(\mathsf{Y} \odot \mathsf{X}^\mathrm{T}) \\
\mathsf{Y}' = \mathrm{FFN}(\mathsf{M}_{\mathsf{YH}}\odot\mathsf{H}+\mathsf{M}_{\mathsf{YX}}\odot\mathsf{X})
\label{equ:parallel}
\end{gather}

Then $\mathsf{Y}'$ will be merged with $\mathsf{Y}$ using the residual connection and layer norm. Notably, we evaluate our model equipped with different cross-granularity architecture and analyse the result in \textit{Ablation Studies}. The results show that the our model with the parallel cross-granularity structure performs better, and we keep this structure in all comparisons with other methods. We apply the cross-granularity attention in all blocks for each branch of transformers. In this way, the visual entities of different granularities can enhance each other with more discriminative semantics for video captioning. It is noteworthy that we can leverage the video-text features in history to obtain the long-term sentence-level recurrence to generate the next sentences according to \citet{DBLP:conf/acl/LeiWSYBB20}, which can further enhance the performance of our model.

\subsection{Optimization Objective}
As we stated before, we introduce specified loss to each stream to guide the training of our COST method. This approach helps in preserving the semantics of visual features at different granularities and facilitate their complementary nature. The loss is formed by three terms, \ie, $\mathcal{L}_{v}(\cdot, \cdot)$ for the Video-Text transformer, $\mathcal{L}_{d}(\cdot, \cdot)$ for the Detection-Text transformer, and $\mathcal{L}_{a}(\cdot, \cdot)$ for the Action-Text transformer, \ie,
\begin{equation}
	\begin{split}
		\mathcal{L} = & \mathcal{L}_{v}(\ell_v, [\{f^t_1, \cdots, f^t_{N_t}, [PAD]_{\times N_2}\}]) \\
		& + \lambda_{d} \cdot \mathcal{L}_{d}(\ell_d, [\{f^d_1, f^d_2, \cdots, f^d_{N_d}\}]) \\
		& + \lambda_{a} \cdot \mathcal{L}_{a}(\ell_a, [\text{CLS}^a])
	\end{split}
\label{equ:loss}
\end{equation}
where $\lambda_{d}$ and $\lambda_{a}$ are the pre-set parameters used to balance these three terms. $\mathcal{L}_{v}(\cdot, \cdot)$ is the cross-entropy loss used to penalize the errors of the predicted captions by the linguistic tokens from Video-Text branch comparing to the ground-truth descriptions $\ell_v$, which are the indexes of words in caption sentences. $\mathcal{L}_{d}(\cdot, \cdot)$ is also the cross-entropy loss used to penalize the errors of the predicted categories of objects by the visual tokens from Detection-Text branch comparing to the pseudo category labels generated by the Faster R-CNN detector\footnote{Notably, we do not use the annotated objects for model training, but use the pseudo category label $\ell_d$ generated by the Faster R-CNN detector as the ground-truth label to enforce the network to maintain the original encoded semantic information of detector. The reason is that the ground-truth of video captioning does not explicitly provide the categories of items in the scene and there is a one-to-one correspondence between pseudo categories and extracted features.}. $\mathcal{L}_{a}(\cdot, \cdot)$ is designed as the multi-label classification loss because there may be multiple actions in a video clip. Specifically, we first aggregate all action tokens as the granularity token $[\text{CLS}^a]$ and then compute the confidence score using a fully-connected layer, \ie, $\mathrm{FC}([\text{CLS}^a])$. Following \citet{DBLP:conf/ijcai/ZhangCXDTCHSC21} and \citet{DBLP:conf/cvpr/SunCZZZWW20}, the loss function is computed as
\begin{equation}
	\begin{split}
		\mathcal{L}_{a}(\ell_a,[\text{CLS}^a]) & = \log \big(1+\sum_{i \in \Omega_\text{pos}(\ell_a)} e^{-s_{i}}\big) 
		 +\log \big(1+\sum_{j \in \Omega_\text{neg}(\ell_a)} e^{s_{j}}\big)
	\end{split}
\end{equation}
which expects the confidence scores $s_i$ of the existing actions $\Omega_\text{pos}(\ell_a)$ in video sequences are greater than the predefined threshold $0$ while the confidence scores $s_j$ of non-existing actions $\Omega_\text{neg}(\ell_a)$ are less than $0$. The ground-truth action labels are the most common verbs in caption sentences, which are retrieved by the off-the-shelf part-of-speech tagger method \citep{DBLP:conf/iccv/SunMV0S19}. 

\section{Experiments}
\subsection{Datasets And Evaluation Metrics}
{\noindent {\bf Datasets.}}
We conduct several experiments on two challenging datasets, \ie, YouCookII \citep{DBLP:conf/aaai/ZhouXC18} and ActivityNet Captions \citep{DBLP:conf/iccv/KrishnaHRFN17}. 

1) \textit{YouCookII} includes $2,000$ long untrimmed videos describing $89$ cooking recipes, where each video contains one reference paragraph and is further split into several event segments with annotated sentences. $1,333$ and $457$ video sequences are used for training and validation, respectively. 

2) \textit{ActivityNet Captions} is a large-scale dataset formed by $10,009$ videos for training and $4,917$ videos for validation and testing. Notably, since the testing set is not publicly available, following \citet{DBLP:conf/cvpr/ZhouKCCR19}, the original validation set is split into the \textit{ae-val} subset with $2,460$ videos for validation and the \textit{ae-test} subset with $2,457$ videos for testing.

3) \textit{MSVD} consists of 1970 video clips collected from YouTube and each clip is spanning over 10 to 25 seconds, with 40 sentences as annotations. Following \citet{DBLP:conf/iccv/0001J21, DBLP:conf/cvpr/YeLQWH022}, we split the dataset into $1,200$, $100$, and $670$ video clips for training, validation and testing, respectively. 

{\noindent {\bf Data Preprocessing.}}
For fair comparisons, we try to keep consistent with the previous works. For YouCookII and ActivityNet Captions datasets, we use the features extracted provided by \citet{DBLP:conf/cvpr/ZhouZCSX18} as video features and motion features to represent videos. Specifically, firstly the video is down-sampled to extract frame every $0.5$s, then the video feature is extracted from the 'Flatten-$673$' layer in ResNet-$200$ \citep{DBLP:conf/cvpr/HeZRS16} with the dimension of $2048$ and motion feature is extracted from the 'global pool' layer in BN-Inception \citep{DBLP:conf/icml/IoffeS15} with the dimension of $1024$. Both networks are pre-trained on ActivityNet dataset \citep{DBLP:conf/cvpr/HeilbronEGN15} for the action recognition task. Following previous works \citep{DBLP:conf/cvpr/YeLQWH022, DBLP:conf/cvpr/ZhangSY0WHZ20}, we utilize InceptionResNetV2 \citep{DBLP:conf/aaai/SzegedyIVA17} and C3D \citep{DBLP:conf/iccv/TranBFTP15} to extract 1536-dim video features and 2048-dim motion features, respectively. The two models are pretrained on ImageNet dataset \citep{DBLP:journals/ijcv/RussakovskyDSKS15} and Sports 1M dataset \cite{DBLP:conf/cvpr/KarpathyTSLSF14} separately. And we use 16-frame clips with 8-frame overlap as input to C3D. For all the three datasets, we uniformly use Faster R-CNN \citep{DBLP:journals/pami/RenHG017} with ResNet-101 as the backbone to extract the object representation as detection features with the dimension of $2048$. Considering the trade-off between accuracy and complexity, we give preference to $5$ detection features with highest confidence and different classes per frame. As we predefine that at most $100$ frames are sampled from one video clips, \ie, $N_v=N_a=102, N_d=502$ with the two special tokens $[\text{CLS}]$ and $[\text{SEP}]$. Meanwhile, we exploit the first $20$ words in the caption sentences and compute the $300$-dim GloVe features, \ie, $N_t=22$ with the two special tokens $[\text{BOS}]$ and $[\text{EOS}]$. It is worth noting that, we selected $20$ as the truncation number firstly to be consistent with the previous method. Concurrently, we also conducted a statistical analysis on the distribution of caption lengths across the three datasets. The results indicate that over $90\%$ of captions fall within this range, and even more than $97\%$ of YouCookII satisfies this criterion. Therefore, this choice is a reasonable decision. For the COOT features, we concatenate the local clip-level ($384$-dim) and the global video-level ($768$-dim) features to describe the videos. After the fully-connected layer, all the tokens are converted into the $768$-dim features.
\begin{table*}[t]
  \centering
  \setlength{\tabcolsep}{10pt}
  \caption{\small Experimental Results on the YouCookII \textit{Val} Subset and ActivityNet Captions \textit{Ae-Test} Subset in the \textbf{Paragraph-Level} Evaluation mode. \textit{COOT} Indicates That the Evaluated Methods Use the Feature Extracted by COOT \citep{DBLP:conf/nips/GingZPB20} Pre-Trained on HowTo100M \citep{DBLP:conf/iccv/MiechZATLS19}. * indicates that VLTinT utilizes extra linguistic features obtained through CLIP \citep{DBLP:conf/icml/RadfordKHRGASAM21}. We Report BLEU@$4$ (B@$4$), METEOR (M), CIDEr-D (C) and Repetition@4 (R@4).}
  	\resizebox{\textwidth}{!}{
    \begin{tabular}{l|c|cccc|cccc}
    \hline
    \multicolumn{1}{c|}{\multirow{2}[0]{*}{Method}} & \multicolumn{1}{c|}{\multirow{2}[0]{*}{\textit{COOT}}} & \multicolumn{4}{c|}{YouCookII (\textit{val})} & \multicolumn{4}{c}{ActivityNet Captions (\textit{ae-test})} \\
       & & \multicolumn{1}{c}{B@4} & \multicolumn{1}{c}{M} & \multicolumn{1}{c}{C} & \multicolumn{1}{c|}{R@4~$\downarrow$} & \multicolumn{1}{c}{B@4} & \multicolumn{1}{c}{M} & \multicolumn{1}{c}{C} & \multicolumn{1}{c}{R@4~$\downarrow$} \\
    \hline
    Vanilla Transformer \citep{DBLP:conf/cvpr/ZhouZCSX18} &\xmark &7.62& 15.65 & 32.26 & 7.83&9.31& 15.54 & 21.33 & 7.45 \\
    Transformer-XL \citep{DBLP:conf/acl/DaiYYCLS19} &\xmark & 6.56  & 14.76 & 26.35 & 6.30 & 10.25 & 14.91 & 21.71 & 8.79 \\
    Transformer-XLRG \citep{DBLP:conf/acl/LeiWSYBB20} &\xmark & 6.63 &14.74 & 25.93 & 6.03  &10.07 & 14.58 & 20.34 & 9.37 \\
    MART \citep{DBLP:conf/acl/LeiWSYBB20} &\xmark & 8.00 & 15.90  & 35.74 & 4.39 & 9.78  & 15.57 & 22.16 & \underline{5.44} \\
    PDVC \citep{DBLP:conf/iccv/WangZLZC021} &\xmark & 7.11 & 15.05 & 26.03 & 7.65 & \underline{11.36} & 15.73 & \underline{25.03} & 10.92 \\
    VLTinT* \citep{DBLP:journals/corr/abs-2211-15103} &\cmark & \underline{9.40} & \textbf{17.94} & \textbf{48.70} & \underline{4.29} & \textbf{14.50} & \textbf{17.97} & \textbf{31.13} & \textbf{4.75} \\
    \hline 
    \hline
    COST &\xmark & \textbf{9.47} & \underline{17.67} & \underline{45.54} & \textbf{4.04} & 11.14 & \underline{15.91} & 24.77 & 5.86 \\
    \hline\hline
    Vanilla Transformer \citep{DBLP:conf/cvpr/ZhouZCSX18}&\cmark & 11.05 & 19.79 & 55.57 & 5.69 & 10.47 & 15.76 & 25.90 & 19.14 \\
    Transformer-XL \citep{DBLP:conf/acl/DaiYYCLS19} &\cmark &-  &-  &-  &-    & 10.57 & 14.76 & 22.04 & 15.85 \\
    MART \citep{DBLP:conf/acl/LeiWSYBB20} &\cmark & 11.30 &\underline{19.85} & 57.24 & \textbf{6.69} & 10.85 & \underline{15.99} & 28.19 & \underline{6.64} \\
    SART \citep{DBLP:journals/tomccap/ManOLSS22} &\cmark & \underline{11.43} & \textbf{19.91} & \underline{57.66} & 8.58 & \underline{11.35} & \textbf{16.21} & \underline{28.35} & 7.18 \\
    COST &\cmark &\textbf{11.56} & 19.67 & \textbf{60.78} & \underline{6.63} & \textbf{11.88} & 15.70 & \textbf{29.64}   & \textbf{6.11}  \\
    \hline
    \end{tabular}}
  \label{tab:main_res}%
\end{table*}%

{\noindent {\bf Evaluation metrics.}}
Similar to former works \citep{DBLP:conf/cvpr/ParkRDR19,DBLP:conf/acl/LeiWSYBB20,DBLP:conf/cvpr/ZhuY20a}, we use several standard metrics to evaluate our method, including BLEU@$n$ (B@$n$) \citep{DBLP:conf/acl/PapineniRWZ02} for $n$-gram precision, METEOR (M) \citep{DBLP:conf/wmt/DenkowskiL14} for $n$-gram with synonym matching, CIDEr-D (C) \citep{DBLP:conf/cvpr/VedantamZP15} for consensus measurement, Rouge(R) \citep{lin2004rouge} for longest subsequence similarities and Repetition@4 (R@4) \citep{DBLP:conf/eccv/XiongDL18,DBLP:conf/cvpr/ParkRDR19} for $n$ repetition in the description. Notably, two evaluation modes are considered, \ie, micro-level and paragraph-level. The micro-level evaluation reports the average score on all video sequences separately; while the paragraph-level evaluation first concatenates the caption sentences of all video sequences and then computes the scores averaged across all videos based on the ground-truth paragraph caption sentences, which aims to preserve the story flow with coherence and conciseness. Following existing works \citep{DBLP:conf/eccv/XiongDL18, DBLP:conf/acl/LeiWSYBB20}, we apply the paragraph-level evaluation to YouCookII and ActivityNet Captions datasets as they annotates multiple clips in each videos. In both modes, CIDEr is used as the primary metric for ranking.

\subsection{Implementation Details}
Our COST algorithm is implemented using PyTorch. All the experiments are conducted on a machine with 2 NVIDIA RTX-3090 GPUs. We train the model using the strategies similar to BERT \citep{DBLP:conf/naacl/DevlinCLT19}. Specifically, we use Adam \citep{DBLP:journals/corr/KingmaB14} with an initial learning rate of $1e-4$, $\beta_1\mbox{=}0.9$, $\beta_2\mbox{=}0.999$, $\normltwo$ weight decay of $0.01$, and the learning rate warmup over the first $2$ epochs. We train the model at most $20$ epochs with early-stop strategy based on CIDEr-D and the batch size is set to $64$. For each branch of transformers, we set the dimension of the feature embeddings $d=768$, the number of transformer blocks $S=2$, and the number of attention heads $h=12$. The loss weights $\lambda_{a}$ and $\lambda_{d}$ in \eqref{equ:loss} are set to $2.0$ and $0.02$ which are verified by multiple experiments. It is noted that text tokens are all initialized as [PAD] at the time of evaluation and the description is generated word by word: one word token is generated at a time and substitutes for [PAD] in the corresponding position, and then sent to the model with previously generated tokens to predict the next one until [EOS] token is generated or the length of sentence reaches the maximum.

\subsection{Evaluation Results}
We compare the proposed COST method with the state-of-the-art methods on the three challenging datasets, \ie, YouCookII, ActivityNet Captions and MSVD. As shown in Table \ref{tab:main_res}, our method obtains the competitive results on both the YouCookII \textit{val} subset and the ActivityNet Captions \textit{ae-test} subset. Without using the COOT features, our method approaches the state-of-art work, \ie, VLTinT \citep{DBLP:journals/corr/abs-2211-15103} on the YouCookII \textit{val} subset. Beside using 3D-CNN network and a human detector to extract global and local visual features, VLTinT utilizes Language-Image Pre-training (CLIP) \citep{DBLP:conf/icml/RadfordKHRGASAM21} to obtain additional linguistic features. And compare our work with other works which don't introduce additional linguistic features, it can be observed that our method improves near $10\%$ CIDEr score compared to the second best method, \ie, MART \citep{DBLP:conf/acl/LeiWSYBB20}.  This is attributed to the fact that we use the multi-branch structure to process visual information at different granularities and make them complementary to each other through the proposed cross-granularity attention module, which make it easier for the model to understand the video content. Besides, our method exploits the local appearance information from the Detection-Text transformer for more accuracy caption generation. We also give experiment results in ablation part to verify their effects respectively. Using the COOT features, the overall video captioning results are significantly improved, which demonstrates that the video features extracted by different pre-trained models have a great impact on the performance of the captioning model, and our COST method also performs favorably against other algorithms by improving over $3\%$ CIDEr score. We observe that the similar trend appears in the ActivityNet Captions \textit{ae-test} subset. It is worth noting that PDVC \citep{DBLP:conf/iccv/WangZLZC021} performs better on ActivityNet but relatively poorly on YouCookII, probably due to the different characteristics of the two datasets: each video of ActivityNet lasts 120s with 3.65 temporally-localized sentences on average while the duration of each video in YouCookII is 320s with 7.7 annotated segments and associated sentences. However, COST has a balanced and satisfactory performance on both datasets, which further proves the effectiveness of our method. 
\begin{table}
  \centering
  \small
  \caption{\small Comparison With the State-of-the-Art Methods on ActivityNet Captions \textit{Ae-Val} Subset in the \textbf{Paragraph-Level} Evaluation Mode. \textit{Det.} And \textit{Re.} Indicate Whether the Model Uses Detection Features And the Sentence-Level Recurrence. * indicates that VLTinT utilizes extra linguistic features obtained through CLIP \citep{DBLP:conf/icml/RadfordKHRGASAM21}.}
  \resizebox{\linewidth}{!}{
    \begin{tabular}{lcccccc}
    \hline
    & \textit{Det.} & \textit{Re.} &  B@4 & M & C & R@4~$\downarrow$ \\ 
	\hline
    \multicolumn{7}{l}{\textbf{LSTM based methods:}}\\
	MFT \citep{DBLP:conf/eccv/XiongDL18} & \xmark & \cmark & 10.29 & 14.73 & 19.12 & 17.71 \\
	HSE \citep{DBLP:conf/eccv/ZhangHS18} & \xmark & \cmark & 9.84 & 13.78 & 18.78 & 13.22 \\
	\hline
	\multicolumn{7}{l}{\textbf{LSTM based methods with detection feature:}} \\
	GVD \citep{DBLP:conf/cvpr/ZhouKCCR19} & \cmark & \xmark & 11.04 & 15.71 & 21.95 & 8.76 \\
	GVDsup \citep{DBLP:conf/cvpr/ZhouKCCR19} & \cmark & \xmark  & 11.30 & 16.41 & 22.94  & 7.04 \\
	AdvInf \citep{DBLP:conf/cvpr/ParkRDR19} & \cmark & \cmark & 10.04 & \underline{16.60} & 20.97 & 5.76 \\
	\hline
	\multicolumn{7}{l}{\textbf{Transformer based methods:}} \\
	Vanilla Transformer \citep{DBLP:conf/cvpr/ZhouZCSX18} & \xmark  & \xmark & 9.75 & 15.64 & 22.16 & 7.79 \\
	Transformer-XL \citep{DBLP:conf/acl/DaiYYCLS19} & \xmark & \cmark & 10.39 & 15.09 & 21.67 & 8.54  \\
	Transformer-XLRG \citep{DBLP:conf/acl/LeiWSYBB20} & \xmark  & \cmark & 10.17 & 14.77 & 20.40 & 8.85   \\
	MART \citep{DBLP:conf/acl/LeiWSYBB20} & \xmark  & \cmark & 10.33  & 15.68 & 23.42 & \underline{5.18} \\
	PDVC \citep{DBLP:conf/iccv/WangZLZC021} & \xmark  & \xmark & \underline{11.80}  & 15.93 & \underline{27.27}  & 10.68  \\
     SART \citep{DBLP:journals/tomccap/ManOLSS22} &\xmark  & \cmark & 11.35 & 16.21 & 28.35 & 7.18  \\
    VLTinT* \citep{DBLP:journals/corr/abs-2211-15103} & \xmark  & \xmark & \textbf{14.93} & \textbf{18.16} & \textbf{33.07} & \textbf{4.87} \\
    \hline
	COST & \cmark  & \cmark & 11.22  & 16.58 & 25.70 & 7.09 \\
	\hline
	\end{tabular}}
\label{tab:anet_val_res}
\end{table}
As presented in Table \ref{tab:anet_val_res}, we also compare the proposed method to some LSTM based methods with input detection features on the ActivityNet Captions \textit{ae-val} subset. Note that Table \ref{tab:main_res} and \ref{tab:anet_val_res} don't list same number of compared methods is because these LSTM-based methods had only evaluated on the \textit{ae-val} subset of Activity Captions and provided the corresponding results. Compared to AdvInf \citep{DBLP:conf/cvpr/ParkRDR19}, which also inputs multi-modal features (concatenation of image recognition, action recognition and object detection features) and designs three discriminators to enhance the fluency and relevance in the generated captions, our method produces higher scores for both B@4 and CIDEr, demonstrating the superiority of the collaborative transformers to learn multi-modal representations over LSTM. Meanwhile, MART \citep{DBLP:conf/acl/LeiWSYBB20} without input detections performs inferior than our method in terms of B@4, M and C metrics, even though it utilizes complex memory module to store video-text features in history to generate more coherent and accurate captions. It indicates that detection features are important entity for video captioning task which is neglected by recent transform-based approaches, and aligning the interactions between these visual features of different granularities can contribute greatly to generating accurate captions. PDVC \citep{DBLP:conf/iccv/WangZLZC021} adds multiple temporal convolutional layers to obtain feature sequences across multiple resolutions as input and performs better than our method in terms of B@4 and C metrics, but the sentences generated by it have significant redundancy (with high R@4 metric) due to complex input. It is worth noting that both SART \citep{DBLP:journals/tomccap/ManOLSS22} and VLTinT \citep{DBLP:journals/corr/abs-2211-15103} both performs well on this subset while based on different solutions: the former one proposes a scenario understanding module to enhance the corresponding ability of their model while the latter captures coherent semantics by introducing more powerful linguistic features, which provides insights for future exploration in this dataset.

\begin{table}
 \centering
  \small
  \setlength{\tabcolsep}{2pt}
\caption{\small Comparison With State-of-the-Art Methods on YouCookII \textit{Val} Subset in the \textbf{Micro-Level} Evaluation Mode. * indicates the evaluated method had been pre-trained on large-scale datasets and $\dag$ indicates the evaluate method using multi-modal features extracted by COOT \citep{DBLP:conf/nips/GingZPB20}.}
\begin{tabular}{@{}c|ccccc @{}}
\hline
Method & B@3 & B@4 & M & R & C  \\
\hline
Masked Trans. \citep{DBLP:conf/cvpr/ZhouZCSX18} & 7.53 & 3.84 & 11.55 & 27.44 & 0.38 \\
S3D \citep{DBLP:journals/corr/abs-1712-04851} & 6.12 & 3.24 & 9.52 & 26.09 & 0.31 \\
VideoBERT* \citep{DBLP:conf/iccv/SunMV0S19} & 6.80 & 4.04 & 11.01 & 27.50 & 0.49\\
VideoBERT+S3D* \citep{DBLP:conf/iccv/SunMV0S19} & 7.59 & 4.33 & 11.94 & 28.80 & 0.50\\
ActBERT* \citep{DBLP:conf/cvpr/ZhuY20a} & 8.66 & 5.41 & 14.30 & 30.56 & 0.65\\
SwinBERT* \citep{DBLP:journals/corr/abs-2111-13196} & \textbf{13.80} & \textbf{9.00} & \underline{15.60} & \textbf{37.30} & \textbf{1.09} \\
\hline
COST & 10.69 & 6.63 & 12.61 & 31.09 & 0.71 \\
$\text{COST}^\dag$ & \underline{13.50} & \underline{8.65} & \textbf{15.62} & \underline{36.53} & \underline{1.05}\\
\hline
\end{tabular}
\label{tab:yc2_res_micro_level}
\end{table}

\begin{figure*}
\centering
\includegraphics[width=0.93\linewidth]{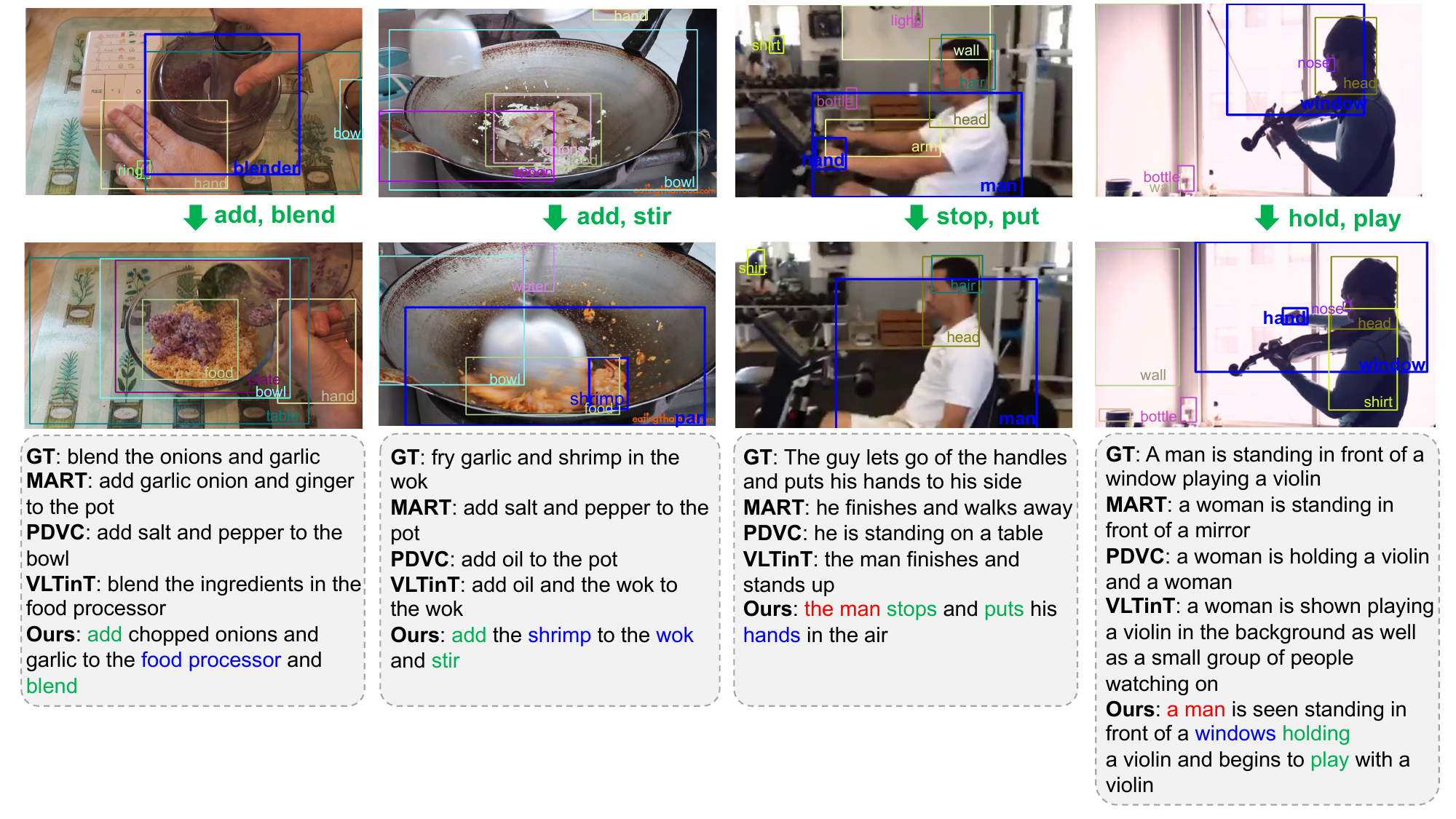}
\caption{\small Qualitative comparison with MART \citep{DBLP:conf/acl/LeiWSYBB20}, PDVC \citep{DBLP:conf/iccv/WangZLZC021} and VLTinT \citep{DBLP:journals/corr/abs-2211-15103} on YouCookII \textit{val} and ActivityNet Captions \textit{ae-val} split. The Subject, Predicate, and Object in a sentence are highlighted in the \textcolor{red}{red}, \textcolor{green}{green} and \textcolor{blue}{blue} fonts, respectively. Best viewed in color.}
\label{fig:visual_examples}
\end{figure*}
According to Table \ref{tab:yc2_res_micro_level}, our method outperforms several BERT based methods which pre-trained on large-scale datasets in terms of the micro-level evaluation. In particular, ActBERT \citep{DBLP:conf/cvpr/ZhuY20a} pre-trains on HowTo100M \citep{DBLP:conf/iccv/MiechZATLS19} firstly and then relies on better visual modalities than our method (for example, its extraction network for action feature pre-trained on Kinetics dataset \citep{DBLP:journals/corr/KayCSZHVVGBNSZ17}). However, it directly focuses on the learning the alignment between text and other visual modalities, which is difficult to exploit the discriminative semantic information. In contrast, our cross-granularity attention module learns from three visual-linguistic interactions of different granularities using the three-stream transformers, producing better CIDEr score. SwinBERT \citep{DBLP:journals/corr/abs-2111-13196} is proposed as a pure Transformer-based end-to-end architecture for video captioning, which leverages VidSwin \citep{DBLP:journals/corr/abs-2106-13230} pre-trained on Kinetics-600 to extract spatial-temporal video representations and then proposes Multi-modal Transformer Encoder takes them as input to generate a natural sentence, and achieves significant improvements compared to the previous methods. In order to make a relatively fair comparison as we had not pre-trained our model on large-scale datasets, such as HowTo100M or Kinectics, we use the visual features provided by COOT \citep{DBLP:conf/nips/GingZPB20} as input, whose video embedding network pre-trained on HowTo100M, and obtain results comparable to SwinBERT, which demonstrates the potential of our method.
\begin{figure*}[htb]
\centering
\includegraphics[width=0.9\linewidth]{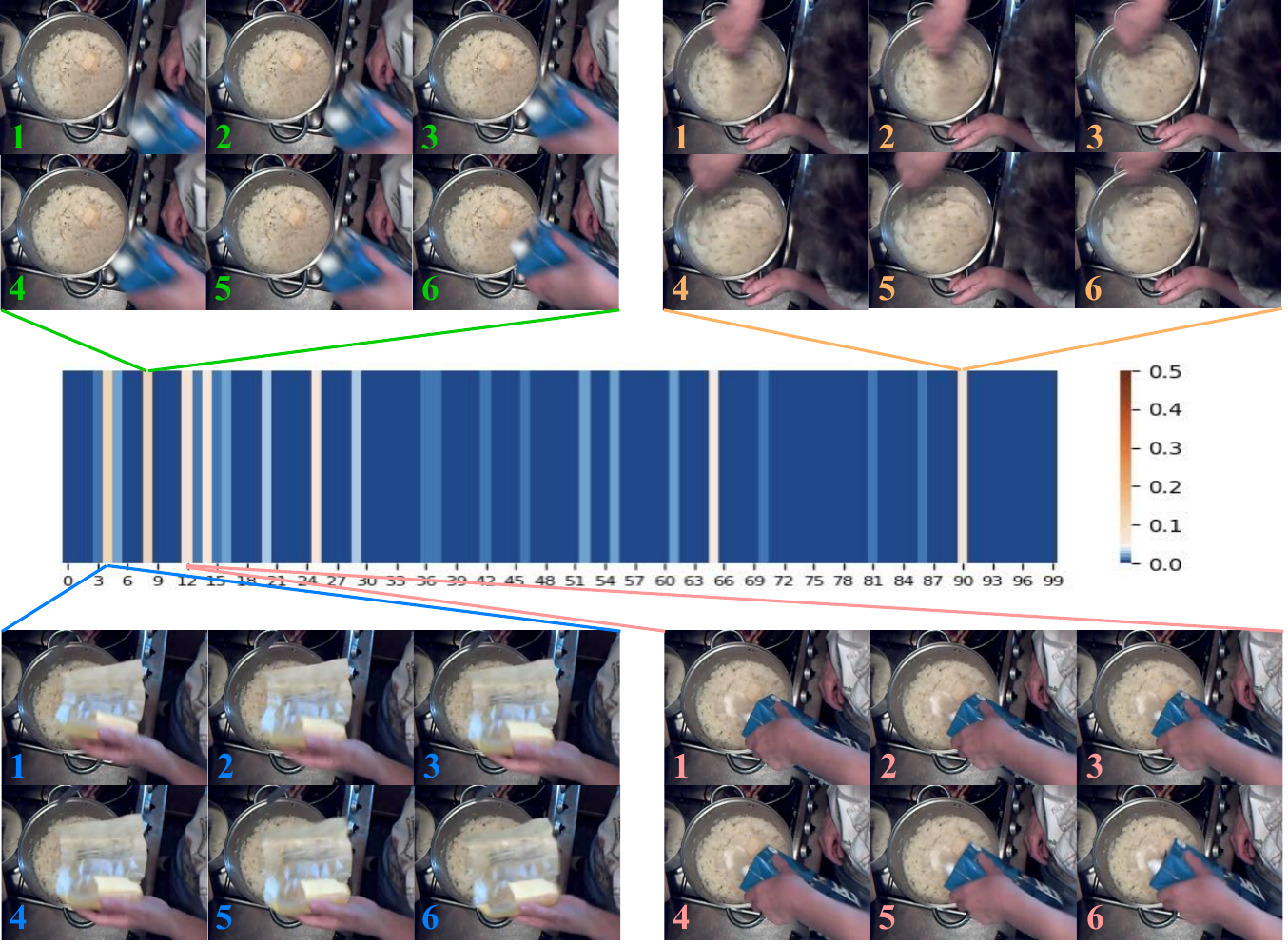}
\caption{\small Visualization of attention score map of all action tokens to $[\text{CLS}^a]$ token in Action-Text transformer for given video clips with ground-truth caption "\textit{add butter and milk to the mashed potatoes and mix}". More crimson color indicates high attention score. We also show the video segments corresponding to the top four attention values. The number in the lower left corner of each picture indicates the time order. Best viewed in color.}
\label{fig:act_attention}
\end{figure*}

\begin{table}
 \centering
  \small
  \setlength{\tabcolsep}{4pt}
\caption{\small Comparison With State-of-the-Art Methods on MSVD \textit{Test} Subset in the \textbf{Micro-Level} Evaluation Mode.}
\begin{tabular}{@{}c|ccccc @{}}
\hline
Method & B & M & R & C  \\
\hline
OA-BTG \citep{DBLP:conf/cvpr/ZhangP19a} & \underline{56.9} & 36.2 & - & 90.6 \\
MSGA \citep{DBLP:conf/aaai/ChenJ19} & 52.5 & 34.1 & 71.3 & 88.7 \\
ORG-TRL \citep{DBLP:conf/cvpr/ZhangSY0WHZ20} & 54.3 & 36.4 & 73.9 & 95.2 \\
JCRR \citep{DBLP:conf/aaai/HouWZQJL20} & \textbf{57.0} & 36.8 & - & 96.8 \\
SGN \citep{DBLP:conf/aaai/RyuKKY21} & 52.8 & 35.5 & 72.9 & 94.3 \\
MGRMP \citep{DBLP:conf/iccv/0001J21} & 55.8 & \underline{36.9} & \textbf{74.5} & \underline{98.5} \\
BFSD \citep{DBLP:journals/corr/abs-2211-15076} & 51.2 & 35.7 & 72.9 & 96.7 \\
\hline
COST & 56.8 & \textbf{37.2} & \underline{74.3} & \textbf{99.2} \\
\hline
\end{tabular}
\label{tab:msvd_res}
\end{table}

We also compare our methods with the state-of-art methods on MSVD dataset, and the results are presented in Table \ref{tab:msvd_res}. As shown in Table \ref{tab:msvd_res}, our COST could approach or surpass the state-of-art method. Considering the significant differences between the characteristics of this dataset and the former two datasets, for example, the duration of video and the umber of annotations corresponding to each video, which further proves the robustness of our method.

Furthermore, the qualitative results of MART \citep{DBLP:conf/acl/LeiWSYBB20}, PDVC \citep{DBLP:conf/iccv/WangZLZC021}, VLTinT \citep{DBLP:journals/corr/abs-2211-15103} and our COST are shown in Fig. \ref{fig:visual_examples}. From Fig. \ref{fig:visual_examples}, it is evident that our COST approach outperforms the compared methods in terms of generating accurate and concise captions, regardless of the complexity of the scenarios. This is attributed to two reasons. First, using the Detection-Text transformer, the Objects in caption sentences can be learned explicitly (\eg, \textit{shrimp}, \textit{hand}, and \textit{windows}) or implicitly (\eg, from \textit{bowl} to \textit{food processor}, and from \textit{pan} to \textit{wok}). Second, the Action-Text transformer in our method can perceive the key actions in caption sentences such as \textit{add}, \textit{blend}, \textit{put} and \textit{hold}. In comparison, the existing methods often fail to recognize these crucial subjects, objects, and verbs. The results indicate that under the constraint of our objective, these two branches of transformers can enforce the network to learn the key elements and supply visual-linguistic interactions with discriminative semantics for captions generation.

Lastly, as there exists many videos shot by moving cameras in our evaluated datasets, for example, YouCookII contains 2000 long untrimmed videos describing 89 cooking recipes, and many videos are shot with varying perspectives for providing good viewing experience, and we wonder whether the performance of our model decreases significantly under such moving camera situation. We have random choose several video clips shot by apparent moving cameras without cherry pick and evaluate our COST on it, and the results are shown on \ref{tab:evaluation_on_moving_scenes}. It can be observed that there is no significant decrease in evaluation metrics, which further confirms the robustness of our method.

\begin{table}[!ht]
    \small
    \setlength{\tabcolsep}{2pt}
    \centering
    \caption{Evaluation Results on Moving Scenes in YouCookII. * indicates the subset picked from YouCookII shot with apparent camera movements.}
    \begin{tabular}{c|cccc}
    \hline
    Dataset & B@4 & M & C & R@4 $\downarrow$ \\
    \hline
    YouCookII* & 9.21 & 17.43 & 43.61 & 5.62 \\
    YouCookII & 9.47 & 17.67 & 45.54 & 4.04 \\
    \hline
    \end{tabular}
    \label{tab:evaluation_on_moving_scenes}
\end{table}

\begin{figure*}[ht]
\centering
\includegraphics[width=0.94\linewidth]{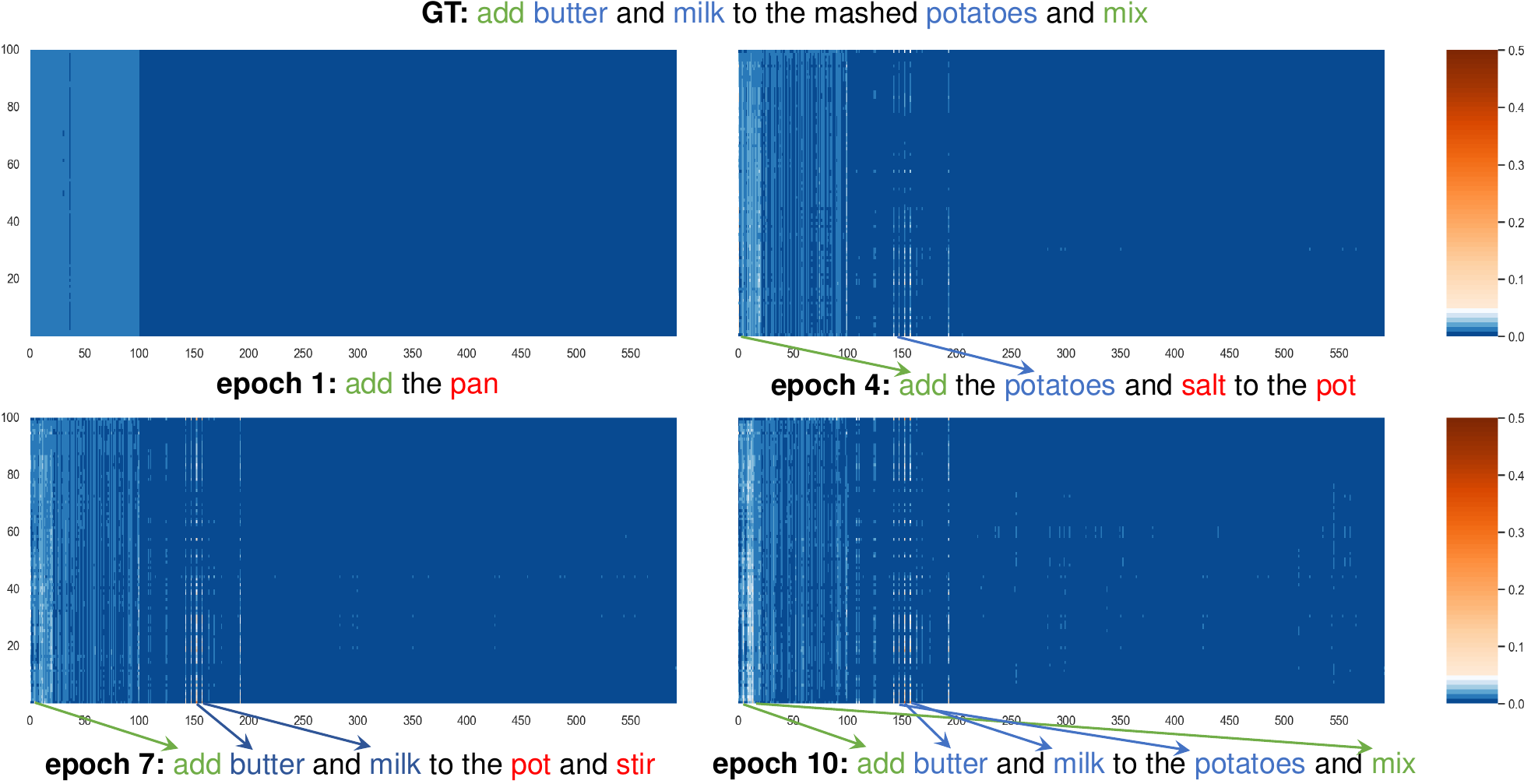}
\caption{\small Heatmap used to indicate the affinity matrix ${\cal M}_{\cal H}$ in the Video-Text transformer, where the row denotes the video tokens and the column denotes the action and detection tokens. The false predictions of nouns and verbs are denoted in red font. For clarity, we only show a few epochs in the training phase. Best viewed in color and zoom in.}
\label{fig:cross_attention}
\end{figure*}

\subsection{Visualization of Self-Attention in Action-Text Transformer}
In Fig. \ref{fig:act_attention}, the central heatmap shows the visualization of the attention scores in self-attention between $[\text{CLS}^a]$ and all action tokens in Action-Text transformer. We obtain the scores by averaging attention scores from all heads of the multi-head self-attention module in our last transformer block. We choose this branch because its granularity token $[\text{CLS}^a]$ is used as the input for action classification, and whether it can utilize action tokens effectively partially reflects the ability of our model to understand the context of video. The ground-truth caption of chosen video is "add butter and milk to the mashed potatoes and mix" and the segments corresponding to top four attention values are presented, where each one consists of 6 figures as setting in motion feature extraction network. According to the chronological order of the action, we can see that the four video segments are: finishing adding butter, readying to pour milk, pouring the milk and stirring the mixture, which proves that our Action-Text Transformer can capture the key actions effectively via self-attention although there exists significant redundancy in the action tokens (vast majority of this video's content is mixing the mixture).

\subsection{Visualization of Cross-Granularity Attention in Video-Text Transformer}

To better understand the cross-granularity attention module, we use the heatmap to visualize the cross-granularity attention scores of the Video-Text transformer in Fig. \ref{fig:cross_attention}. For the convenience of display, we concatenate the affinity matrix $\mathcal{M}_{\mathcal{HY}}$ and $\mathcal{M}_{\mathcal{HX}}$ to get $\mathcal{M}_{\mathcal{H}}$. At epoch $1$, all values in the heatmap are similar and the maximal value is only $0.001$. The corresponding caption results are noisy with the false predictions of nouns and verbs, \eg, \textit{pan} instead of \textit{butter}. It indicates that the cross-granularity attention is randomly initialized and interactions between different modalities are not learned to align well. After training for several epochs, a few entities dominate the heatmap with the maximal value of $0.3$ (see the bright vertical lines in the heatmap). It indicates that our cross-granularity attention module can successfully exploit the most relevant entities from other modalities to inject the information from other branches of transformers. This way, the verb \textit{stir} and the noun \textit{pan} can be corrected to \textit{mix} and \textit{butter} for more accurate caption results.

\begin{table}
  \centering
  \small
  \setlength{\tabcolsep}{9pt}
  \caption{\small Ablation on multi-modality features and three-stream architecture.}
    \begin{tabular}{l|cccc}
    \hline
    \multicolumn{1}{c|}{\multirow{2}[0]{*}{COST Variants}} & \multicolumn{4}{c}{YouCookII (\textit{val})} \\
    \cline{2-5}
       & \multicolumn{1}{c}{B@4} & \multicolumn{1}{c}{M} & \multicolumn{1}{c}{C} & \multicolumn{1}{c}{R@4 ~$\downarrow$} \\
    \hline
    COST-1 (v) & 6.59  & 14.29 & 29.20  & 6.49 \\
    COST-1 (v+a) & 7.72  & 15.45 & 33.98 & 4.60 \\
    COST-1 (v+a+d) & 8.73  & 16.90  & 38.63 & 4.66 \\
    COST-2 (v+a) & 7.79  & 15.74 & 34.99 & 5.82 \\
    COST-2 (v+d) & 9.04  & 17.31 & 43.09 & 4.59 \\
    COST-3 (v+a+d) & \textbf{9.47}  & \textbf{17.67} & \textbf{45.54} & \textbf{4.04} \\
    \hline
    \end{tabular}%
  \label{tab:ablation_yc2}%
\end{table}%

\begin{figure*}
\centering
\includegraphics[width=0.9\linewidth]{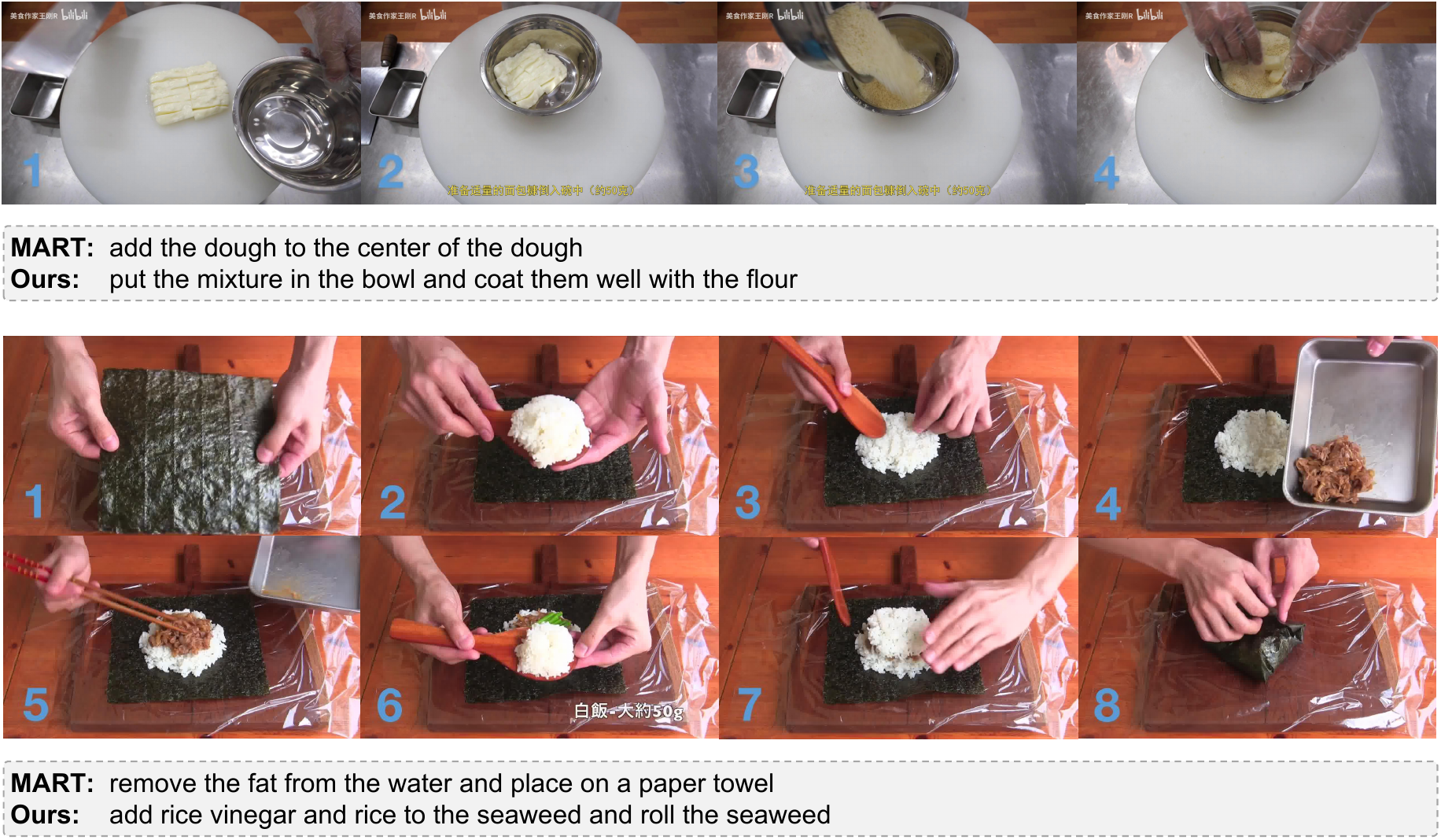}
\caption{\small Generated Descriptions for Internet videos from the state-of-art MART \citep{DBLP:conf/acl/LeiWSYBB20} method and our COST method. The blue number in the lower left corner of each picture indicates the time order.}
\label{fig:generation}
\end{figure*}

\subsection{Ablation Studies}
To study the influence of different components in the proposed method, we conduct detailed ablation study on the YouCookII \textit{val} subset. Notably, we use the appearance features extracted by TSN \citep{DBLP:conf/eccv/WangXW0LTG16} in all COST-$k$ variants, where $k$ denotes the number of transformer branches retained in our COST method.

{\noindent {\bf Effectiveness of multi-modality features.}}
To verify the effectiveness of the multi-modality features, we construct three COST-1 variants, \ie, COST-1 (v), COST-1 (v+a) and COST-1 (v+a+d). In particular, we only use the Video-Text transformer, but change the input features as the combinations of the GloVe text features and the concatenated features from video (v), action (a) and detection (d). As shown in Table \ref{tab:ablation_yc2}, the scores under all metrics are improved considerably by integrating the action or detection features. Moreover, the CIDEr score is boosted from $29.20\%$ to $38.63\%$, if we include all the three-modality features. It indicates that multi-modality features definitely facilitate to generate more accurate video captions. And it can be observed that the incorporation of detection features into COST has yielded notable performance improvements, highlighting the potential for leveraging more detailed features to advance the field of video captioning.

{\noindent {\bf Effectiveness of three-stream transformers.}}
To demonstrate the effectiveness of the three-stream transformers compared to the simple feature concatenation used in COST-1, we construct three COST-$k$ ($k=2,3$) variants, shown in Table \ref{tab:ablation_yc2}. It can be seen that the accuracy can be improved by using the three-stream transformers, \ie, the CIDEr score improved from $38.63\%$ to $45.54\%$. Meanwhile, the accuracy is considerably improved by using more branches of transformers. This is because our cross-granularity attention module is able to capture the most relevant semantic information from different visual-linguistic interactions. Compared to the COST-1 variants, our full model can obtain a more discriminative feature representations for video captioning.

\begin{table}
  \centering
  \small
  \setlength{\tabcolsep}{7pt}
  \caption{\small Ablation on training objectives.}
    \begin{tabular}{l|cccc}
    \hline
    \multicolumn{1}{c|}{\multirow{2}[0]{*}{COST Variants}} & \multicolumn{4}{c}{YouCookII (\textit{val})} \\
    \cline{2-5}
       & \multicolumn{1}{c}{B@4} & \multicolumn{1}{c}{M} & \multicolumn{1}{c}{C} & \multicolumn{1}{c}{R@4 ~$\downarrow$} \\
    \hline
    COST w/o $\mathcal{L}_{a}$ & 8.95  & 17.31 & 42.90  & 5.27 \\
    COST w/o $\mathcal{L}_{d}$ & 9.10  & 16.58 & 41.00  & 5.56 \\
    COST w/o $\mathcal{L}_{d}$ and $\mathcal{L}_{a}$ & 9.01  & 16.29 & 40.58  & 5.91 \\
    \hline
    COST & 9.47 & 17.67 & 45.54 & 4.04 \\ 
    \hline
    \end{tabular}%
  \label{tab:ablation_attention_loss}%
\end{table}%
{\noindent {\bf Effectiveness of training objective.}}
To investigate the effectiveness and rationality of proposed training objective, we do ablation experiments to obtain corresponding results in Table \ref{tab:ablation_attention_loss}. We observe that removing either $\mathcal{L}_{a}$ or $\mathcal{L}_{d}$ has a negative impact on the performance, especially when they are all removed, \ie, 40.58 {\it vs.} 45.54 for CIDEr. This demonstrates that using appropriate objective to make the semantics of visual-linguistic interactions explicit can definitely contribute to accurate captions.

\begin{table}
  \centering
  \small
  \setlength{\tabcolsep}{8pt}
  \caption{\small Ablation on cross-granularity attention module. \textit{concat} and \textit{para} indicate concatenated and parallel cross-granularity attention module separately. COST w/o cgam. denotes a COST variant without cross-granularity attention module.}
    \begin{tabular}{l|cccc}
    \hline
    \multicolumn{1}{c|}{\multirow{2}[0]{*}{COST Variants}} & \multicolumn{4}{c}{YouCookII (\textit{val})} \\
    \cline{2-5}
       & \multicolumn{1}{c}{B@4} & \multicolumn{1}{c}{M} & \multicolumn{1}{c}{C} & \multicolumn{1}{c}{R@4 ~$\downarrow$} \\
    \hline
    COST w/ \textit{concat} & 9.27  & 17.40 & 43.43 & 6.34 \\
    COST w/ \textit{para} & 9.47 & 17.67 & 45.54 & 4.04 \\
    COST w/o cgam. & 7.55  & 15.11 & 30.32 & 8.03 \\
    \hline
    \end{tabular}%
  \label{tab:ablation_cross-ga}%
\end{table}%

{\noindent {\bf Effectiveness of cross-granularity attention module.}}
To verify the importance of cross-granularity attention module to the success of our method, we compare the performance of COST with the two proposed architectures(concatenated and parallel architectures) separately and the variant with cross-granularity attention module removed. As shown in Table \ref{tab:ablation_cross-ga}, our method with cross-granularity module using parallel structure outperforms that with concatenated structure in all evaluation metrics, it is probably because that there exists feature gap between the embeddings from different branches and the softmax operation may hinders the information of one branch being suppressed during fusion. And after removing the cross-granularity module, CIDEr drops significantly, \ie, 30.32 {\it vs.} 45.54, which demonstrate the effectiveness of this module to align the visual-linguistic interactions of different granularities for accurate captions.

\section{Generalization}
In order to verify the generalization and practicality of our method, we collected some videos outside the trained datasets and evaluate the quality of generated descriptions to them. Because the video included in YouCookII dataset has relatively clear action intention and contains abundant clips recording the interactions between human and surrounding objects, we collected some cooking videos from the Internet and applied model trained on YouCookII to them. The way to extract features from collected videos keeps consistent with that to YouCookII, and we adopt trained models on both short and long video clips to test the performance comprehensively. We list the captions generated by MART \citep{DBLP:conf/acl/LeiWSYBB20} and our method in Fig. \ref{fig:generation}. The first clip is one step of making fried milk, which is a dessert of Cantonese cuisine, and lasts about $20$ seconds to coat the fried milk with breadcrumbs. It can be seen that our method not only makes a correct judgment on the action(coat), but also correctly recognizes the state of using substances(both breadcrumbs and flour are powdery objects), which is much better than MART. The second clip is one step of making rice ball and lasts about $1$ minute to add rice, cooked beef and asparagus to seaweed and wrap them. Because there are many added contents, our model makes wrong recognition (the beef is recognized as vinegar because of the similar color) and generates redundant words (rice are added twice in the clip). Nevertheless, the generated description is far more accurate in action and much natural than the description generated by MART. Through these two examples, it illustrates that although our model may make mistakes in generating object words due to the interference of similar shape or color, it can make accurate recognition on the actions of the subject and the structure of generated sentence is very natural, which indicates the effectiveness and practical value of our model.

\section{Limitation}
Although our model has demonstrated satisfactory performance on three benchmark datasets and exhibits strong generalization capabilities, there still exists some limitations. Firstly, in terms of evaluation efficiency and model parameters, we observe certain disadvantages when compared to single-stream methods like MART. Specially, under the same hardware conditions, our model generates captions for approximately 2.47 videos per second, and the duration of each videos is about 5 minutes, whereas MART achieves a higher processing rate of 6.85 videos per second. Furthermore, our model has a larger parameter count of 35.6M compared to MART's 25.5M. Additionally, it is worth noting that our model requires additional data processing steps for extracting region features, which requires additional pre-processing time.

\section{Conclusion}
In this paper, we propose the collaborative three-stream transformers to exploit the interactions of objects, and the actions/relations of objects between different modalities of different granularities in spatial-temporal domain. Meanwhile, with the help of proposed training objective to specify the semantics of visual-linguistic interactions in each branch, the cross-granularity attention module is designed to align the interactions modeled by the three branches of transformers, which contributes to more accurate captions generation. Several experiments conducted on the YouCookII, ActivityNet Captions and MSVD datasets demonstrate the effectiveness of our proposed method, and we also evaluate its generalization ability through testing on Internet videos. In the future, we intent to further improve the performance of our model with the help of pre-training on large-scale datasets.

\section*{Acknowledgments}
This work was supported by the Key Research Program of Frontier Sciences, CAS, Grant No. ZDBS-LY-JSC038 for funding this work. Libo Zhang was also supported by Youth Innovation Promotion Association, CAS (2020111). Heng Fan and his employer received no financial support for the research, authorship, and/or publication of this article.


\bibliographystyle{model2-names}
\bibliography{main.bib}
\end{document}